# The Compressed Model of Residual CNDS

Hussam Qassim, David Feinzimer, and Abhishek Verma
Department of Computer Science
California State University
Fullerton, California 92834
Email: {hualkassam, dfeinzimer}(at)csu.fullerton.edu, averma(at)fullerton.edu

*Abstract*— Convolutional neural networks have achieved a great success in the recent years. Although, the way to maximize the performance of the convolutional neural networks still in the beginning. Furthermore, the optimization of the size and the time that need to train the convolutional neural networks is very far away from reaching the researcher's ambition. In this paper, we proposed a new convolutional neural network that combined several techniques to boost the optimization of the convolutional neural network in the aspects of speed and size. As we used our previous model Residual-CNDS (ResCNDS), which solved the problems of slower convergence, overfitting, and degradation, and compressed it. The outcome model called Residual-Squeeze-CNDS (ResSquCNDS), which we demonstrated on our sold technique to add residual learning and our model of compressing the convolutional neural networks. Our model of compressing adapted from the SQUEEZENET model, but our model is more generalizable, which can be applied almost to any neural network model, and fully integrated into the residual learning, which addresses the problem of the degradation very successfully. Our proposed model trained on very large-scale MIT Places365-Standard scene datasets, which backing our hypothesis that the new compressed model inherited the best of the previous ResCNDS8 model, and almost get the same accuracy in the validation Top-1 and Top-5 with 87.64% smaller in size and 13.33% faster in the training time.

## I. INTRODUCTION

The ImageNet Challenge (ILSVRC) [1] is one of the main experiment environment for computer vision and pattern recognition competitions. Convolutional neural networks (CNNs) have accomplished significant breakthroughs in this contest [4] as well in other image classification algorithms [5, 6]. CNN layers learn all the images' scale characteristics and classify [7] in completely framework. The type of characteristics' levels could be elevated by the number of layers utilized in the network. In the ImageNet Challenge (ILSVRC) [1], it was uncovered that the CNNs accuracy can be enhanced by growing the network depth [8, 9]. This demonstrated that the depth of the CNN is of decisive importance. Top results acquired in [8-11] all utilize very deep CNNs exemplary on ImageNet dataset [1]. The advantages of extremely deep models can be expanded by ordinary image classification jobs to other considerable recognition competitions, for example, the object detection and segmentation [13-17]. Further, growing the depth of the CNN by stacking more layers boosts the total no. of parameters, which makes the convergence of back-propagation algorithm extremely slow and apt to overfitting. Furthermore, rising the depth makes the gradients exposed to the problem of demising or exploding of gradients [18, 19].

Utilizing the pre-trained weights of superficial models to set the weight of deeper CNNs was suggested by Simonyan and Zisserman [6]. The suggested method is to resolve the issue of overfitting and slower convergence. Still, utilizing this mechanism to train various mo0dles of gradually depth is calculatingly costly and may cause in hardness of tuning the parameters. Another method to outdo this defy is suggested by Szegedy et al. [9]. Where they utilized auxiliary branches linked to the intermediate layers. These auxiliary branches are classifiers. The main idea of Szegedy et al. [9] of utilizing these classifiers is to raise the gradients to propagate back through layers of the deep NN architecture. Further, the branches are utilized to drive feature maps in the superficial layers to expect the labels utilized at the final layer. Nonetheless, they did not assign a way that can locate the position of where to add these auxiliary classifiers or how to add them. Lee et al. [18] go after similar notion by suggesting to add the auxiliary branches after each middle layer. The losses from these auxiliary classifiers are gathered with the loss of the last layer. This method demonstrated a boosting in the average of convergence. Nonetheless, they did not search the deeply supervised networks (DSN) [18] with extremely deep networks.

Wang et al. [19] proposed CNNs with deep supervision (CNDS). They handled the problem of where to add the auxiliary classifiers. They examined the problem of vanishing gradients in very deep networks to locate which middle layer demands to have an auxiliary classifier. Attaching auxiliary classifiers handles the issue of slower convergence and overfitting. Although the network is now capable of beginning converging, there is another challenging problem, which is the degradation problem. When the depth of the NN raises, the degradation problem raises in deeper networks. Degradation problem starts to douse the accuracy of the network and makes the cause of the degradation issue. Degradation drives to higher training fault as announced in [23, 24] when expansion the network depth by stacking more layers. Furthermore, the degradation that occurs to the accuracy during the training stage demonstrates that various NN models are not likewise easy to

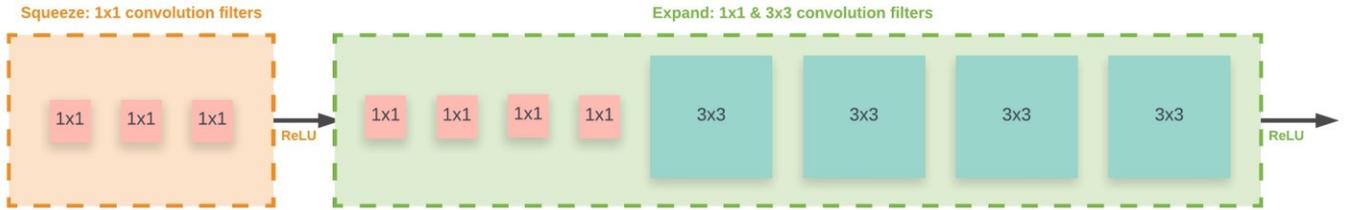

FIGURE 1

Fire module architectural, illustrated the convolution filters in the Fire module. In this figure, $s_{1x1}$ = 3, $e_{1x1}$ = 4, and $e_{3x3}$ = 4. As we demonstrated the convolution filters without the activations. (The SqueezeNet paper [40] influenced the pattern of this figure) the optimize. Residual learning [22] is a state-of-the-art technique that resolves the problem of degradation. In our previous paper, we handle the problem of slower overfitting, convergence, and degradation simultaneously by gathering the CNDS network with residual learning. We attach residual connections [22] to the CNDS [19] eight layers' structure. Experimental results on our paper Residual-CNDS [45] structure demonstrates the benefits of gathering the two structures as it boosts the accuracy of the original CNDS.

A lot of the current research on deep CNNs has concentrated on raising the accuracy on computer vision datasets. For a certain accuracy level, there commonly occur various convolution neural network structures that fulfill the accuracy level. Give an equivalent resolution, a convolution neural network structure with little parameters has several features:

- More effective dole out training. The connection between servers is the shorten agent to the scalability of dole out convolution neural network training. For dole out data-parallel training, connection above is directly commensurate with the number of parameters in the network [36]. In sum, small networks train faster because it is demand less connection.
- Less above when shipping new networks to a customer. For self-driving, companies like Tesla annually install new networks from their servers to client's vehicles. This exercise often points to as an over-the-air update. Client's statements have found that the safety of Tesla's Autopilot semi-self-driving operation has incrementally enhanced with current over-the-air updates [37]. However, over-the-air updates of today's ideal CNN/DNN networks can demand massive data transfers. Utilizing AlexNet [38] for example, this would demand 240MB of communication from the server to the vehicle. On the other hand, smaller networks demand less communication, making recurrent updates more practical.
- Practical FPGA and embedded distribution. FPGAs often have fewer than 10MB of on-chip memory and no off-chip memory or storage. For assumption, an adequately small network can be stored directly on the FPGA instead of being a barrier by memory high frequency [39], while video frames stream over the FPGA in actual time. Moreover, when expanding convolutional neural networks on Application-Specific Integrated Circuits (ASICs), an adequately small network can be stored directly on-chip, and smaller networks could facilitate the ASIC to suit on a smaller die.

From above you can touch, there are certain features of smaller convolutional neural network structures. Consider this, we concentrate straight on the issue of distinguishing a convolutional neural structure with fewer parameters but equal accuracy matched to our previous model ResCNDS [45]. We have discovered such an architecture, which we call Residual-Squeeze-CNDS. In addition, we present our attempt at a more disciplined approach to searching the design space for novel CNN architectures. The new model ResSquCNDS is a very advanced model, which inherited all the features from the previous model ResCNDS [45]. Furthermore, the ResSquCNDS model is smaller in size and faster than the previous model.

In this paper, we have shown our state of the art technique to attach the residual learning connections to the compressed model of the CNDS. Which prevent the degradation problem from occurring to the compressed model of the CNDS, as in the original model. Our new model of compressing also shows a great optimization in term of the size and time. Moreover, our adaptable compression method from the Iandola et al. [40] paper is surpassing the original method in term of generalization and prone to degradation problem, which makes us very confident that our model of compression can be applied to almost every convolutional neural network without any modification.

The remainder of this paper is organized as follows. In section II, we give a brief background of the CNDS network, residual learning, and SqueezeNet. We discuss the details of our proposed Residual-Squeeze-CNDS method in section III. In Section IV, we present the details of very large-scale MIT Places365-Standard scene dataset used in our experiments.

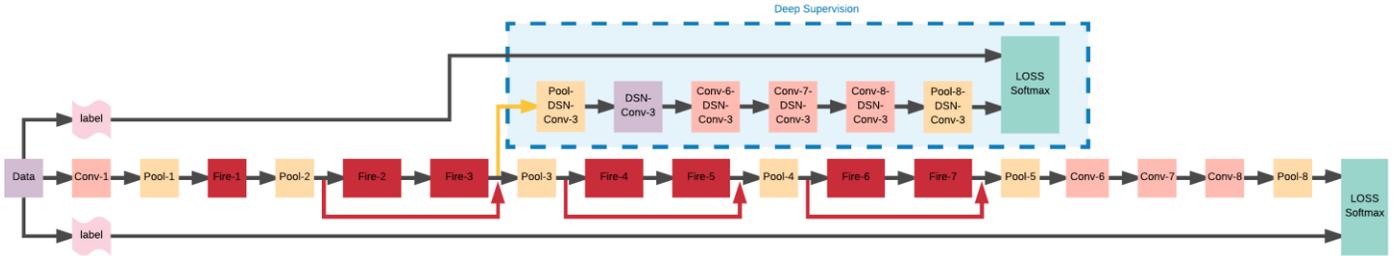

FIGURE 3

THE ARCHITECTURE OF RESIDUAL-SQUEEZE-CNDS. WHICH FIRE MODULES AND RESIDUAL CONNECTIONS ARE SHOWN IN RED

Section V presents our experimental approach. The discussion of the results is given in Section VI. We conclude the paper and suggest future work in section VII.

## II. BACKGROUND

Ever since ILSVRC 2014 [1], the idea to use very deep artificial neural networks have been recognized as very important. Due to this, in recent times, progress has been made on finding efficient ways to train very deep neural networks. Part (A) will outline a CNDS network structure and the ways in which their respective authors utilized vanishing gradients in deciding appropriate locations for auxiliary branch placement. Part (B) will explore the intricacies of a residual learning mechanism. Part (C) contains a demonstration of SqueezeNet [40] and the operations behind it. The following overview and discussion of the CNDS network structure, residual learning mechanism and SqueezeNet [40] will provide a detailed enough description to constitute a good basis for understanding the Residual-Squeeze-CNDS that we aim to outline in this paper.

### A. CNDS Network

Adding auxiliary classifiers which provide further supervision in the training stage improves the generalization of neural networks. Szegedy et al. [7] first proposed this idea through adding subsidiary classifiers which link to middle layers. Despite this contribution, unfortunately, Szegedy et al. [7] did not provide documentation on the location and depth of where to add subsidiary classifiers. Thankfully, Lee et al. [18] provided this information with the introduction of Deeply-Supervised Nets (DSN) in which a support vector machine classifier is connected to the output of each hidden layer in a network. From utilizing this method in training, Lee et al. [18] achieved improvements in the sum of the output layer's loss and subsidiary classifiers losses.

Wang et al. [19] clarified where exactly to add auxiliary classifiers. Deep supervision networks such as the ones proposed by Wang et al. [19] have major distinctions from those proposed by Lee et al [18]. Lee et al. [18] connect the branch classifier in every hidden layer rather than utilizing a gradient focused heuristic in determining when to add an auxiliary classifier. Additionally, Wang et al. [19] implement a small artificial neural network in subsidiary supervision classification. This small network contains one convolutional layer, a small grouping of fully connected layers and one Softmax layer which highly resembles a design introduced by Szegedy et al. [7]. In contrast, Lee et al. [18] used SVM classifiers linked to the outputs of all hidden layers.

To decide where to add auxiliary supervision branches, Wang et al. [19] used the vanishing gradients process. Wang et al. [19] built the neural network while foregoing the use of supervision classifiers. Weights for this network were adapted from the Gaussian pattern, a mean of zero, std set to point-zero-one (0.01) and bias set to zero. Then, Wang et al. [19] would perform between ten and fifty back-propagation epochs in which the mean gradient amount of the shallower layers would be controlled by plotting subsidiary supervision classifiers whenever a mean gradient rate would drop under a certain threshold, such as $10^{-7}$ for instance. In Wang et al. [19] after an average gradient drops under an appointed threshold, the auxiliary classifier was appropriately added.

### B. Residual Learning

Degradation decays optimization in deep convolutional neural networks whereas rising depth should always increase accuracy. Additionally, the error from deeper convolutional neural networks is often higher when compared to that of equivalent superficial neural networks. Nonetheless, He at al. [22] proposed a design with a solution to degradation. In this design, He et al. [22] allowed every few stacked layers to qualify the residual mapping whereas degradation stops layers to fit a required subsidiary mapping. To do this subsidiary mapping formula resembles (2) as opposed to formula (1). He et al. [22] assumed it to be harder to optimize a primary mapping then a residual one.

$$F(x) = H(x) \quad (1)$$
$$F(x) = H(x) - x \quad (2)$$
$$F(x) = H(x) + x \quad (3)$$

A shortcut connection is a process in which one or more layers of a convolutional neural network are passed up [23-25]. A shortcut link can be expressed by formula (3) [22]. He et al. [22] use the idea of shortcut connections in order to perform identity mapping. Shortcut connection output is combined with the output from the stacked layers. An advantage of shortcut connections is that they remain parameter free and only attach trivial numbers for computation operations. Highway networks [21] have shown differences as a result of using shortcut connections in a combination with gating functions with parameters [26]. Another advantage of shortcut connections of the type proposed by He et al. [22] is that they can be optimized through stochastic gradient descent (SGD). Finally, identity shortcut connections are easily implemented through open deep learning libraries [27-30].

*C. SqueezeNet*

Neural network architectures (including those of deep and convolutional denominations) leave a lot of room for choosing different options such as micro/macro architectures, solvers, and additional hyperparameters. Consequently, a good amount of work has been concentrated around designing automated ways for creating neural network architectures with a high level of accuracy. Well, known automated approaches include Bayesian optimization [41], simulated annealing [42], randomized search [43], and genetic algorithms [44]. Each of these approaches has achieved higher accuracy over their respective baseline.

The objective of the SqueezeNet, which proposed by Iandola et al. [40], is to highlight CNN architectures with a small number of parameters and competitive accuracy. The Iandola et al. [40] follow three main strategies in designing CNN architectures:

- Replace 3x3 filters with 1x1 filters.
- Decrease the number of input channels to 3x3 filters.
- Downsample late in the network to give convolution layers' large activation maps.

One and two decrease the number of parameters in CNN while maintaining accuracy. Three will help to maximize accuracy when working with a limited amount of parameters.

### III. PROPOSED RESIDUAL-SQUEEZE-CNDS NETWORK ARCHITECTURE

Our proposed Residual-Squeeze-CNDS contains seven fire modules [40] and four convolution neural layers in the main branch. We attach a Scale layer to all fire modules and the first convolutional layer in the main branch. We assign a stride of two and a size of 3x3 to the kernel in layer one. We replace the second convolutional layer of the ResCNDS [45] with one fire module [40]. We choose to do this because the firing module [40] has 9x fewer parameters than a 3x3 filter counterpart. Additionally, we reduce input channels to only 3x3 filters. The number of parameters in the fire module is the number of input channels multiplied by the number of filters. Therefore, to maintain a few parameters in a CNN architecture it is important to decrease the number of filters and the number of input channels. We give all Max-Pooling layers a kernel size of 3x3 following the idea of the third strategy [40] in that we Downsample late in the network to give convolution layers' large activation maps. A convolution layer will produce an output activation map with a spatial resolution of at least 1x1, but often much larger. The resulting height and width of these activation maps are controlled by two factors: size of the input data and the choice of layers in which downsampling will occur. Downsampling has been implemented into CNN architectures by applying a stride greater than one to some convolution or pooling layers (e.g. (Szegedy et al. [9]; Simonyan & Zisserman [6]; Krizhevsky et al. [2])). When early layers have large stride parameters, as a result, most layers will have small activation maps. However, if most layers have a stride of one and those layers are towards the end of a network, many layers will have large activation map as a result. Our idea is to claim that large activation maps lead to increased classification accuracy. After He & Sun [46] conducted delayed downsampling tests on four unique CNN architectures they observed higher classification accuracy in each result.

In our Residual-Squeeze-CNDS model, we utilize the fire model first proposed in Iandola et al. [40]. The fire module [40] is the composition of a squeeze convolution layer (with 1 x 1 filters) which is then fed into an expand layer comprised of a mix of 1x1 and 3x3 convolution filters. The fire module is illustrated in Figure 1. A fire module has three adjustable dimensions: $s_{1*1}$, $e_{1*1}$, and $e_{3*3}$ [40]. $s_{1*1}$ [40] represents the number of 1x1 filters in the squeeze layer. $e_{1*1}$ [40] represents the number of 1x1 filters in the expanded layer. $e_{3*3}$ [40] is the number of 3x3 filters in the expanded layer. In order to limit the number of input channels to the 3x3 filters, we set $s_{1*1}$ [40] to be less than ($e_{1*1} + e_{3*3}$) [40] as appeared in Table 1.

TABLE (1)

RESIDUAL-SQUEEZE-CNDS MAIN BRANCH ARCHITECTURAL DIMENSIONS
(THE PATTERN OF THIS TABLE WAS INFLUENCED BY THE
SQUEEZENET PAPER [40])

| Layer name/type | s1x1 (#1x1 squeeze) | e1x1 (#1x1 expand) | e3x3 (#3x3 expand) |
|---|---|---|---|
| Input image | - | - | - |
| Conv1 | - | - | - |
| Maxpool1 | - | - | - |
| Fire1 | 16 | 64 | 64 |
| Maxpool2 | - | - | - |
| Fire2 | 32 | 128 | 128 |
| Fire3 | 32 | 128 | 128 |
| Maxpool3 | - | - | - |
| Fire4 | 64 | 256 | 256 |
| Fire5 | 64 | 256 | 256 |
| Maxpool4 | - | - | - |
| Fire6 | 64 | 256 | 256 |
| Fire7 | 64 | 256 | 256 |
| Maxpool5 | - | - | - |
| Conv6 | - | - | - |
| Conv7 | - | - | - |
| Conv8 | - | - | - |
| Average pool8 | - | - | - |

Note: Data not available is marked as '- '

As prescribed by Wang et al. [19], the subsidiary branch, which contains the supervision classifier, follows after the convolutional neural layer which experiences the problem of vanishing gradients (Fire3) as shown in Figure 3. Characteristic maps that the shallower layers create are noisy and it is very important to minimize this noise in the convolutional layers before it reaches the classifiers. In order to minimize the noise, we decrease the dimensionality of the characteristic maps as in Wang et al. [19] paper and then pass them into non-linear functions before placing them into them classifiers. This results in the subsidiary classifier launching with an average pooling layer of a kernel of size 5x5 and a stride of two. Furthermore, a convolutional layer follows the average pooling layer with a kernel of size one and a stride of one, which we add a Scale layer to it. Then we add two additional convolutional layers, rather than fully connected layers, each with a size of 512 and a kernel of size 3x3, connected by a 0.5 dropout ratio. The master and subsidiary branch have their own output convolutional layer with an output that resembles a number of classes in the dataset, a kernel of size one, an average pooling layer and a softmax layer to for classification.

$$W_{main} = (W_1, ..., W_{11}) \qquad (4)$$
$$W_{branch} = (W_{s5}, ..., W_{s8}) \qquad (5)$$

Weights in master branch names are illustrated in formula (4) [19]. These weights match with the eight convolutional layers and three fully connected layers, which resemble those of the original Residual-CNDS model [45]. So the eight convolutional layers equal to the one convolutional layer and seven fire modules in the Residual-Squeeze-CNDS, which each fire module (three convolutional layers) replace one convolutional layer in the original Residual-CNDS [45]. Hence, the three convolutional layers replace the three fully connected layers. Additionally, the auxiliary classifier's weight calculates in formula (5) [19], as the weights in the formula matched to the four convolutional layers in the auxiliary branch. If we contemplate the characteristic map generated from the output layer in the master branch, in the beginning, to be $X_{11}$ then we are able to calculate likelihood by using the softmax function from the labels k =1, ..., K, illustrated by formula (6) [19]. We can calculate the reply through formula (7) [19] if the characteristic map is $S_8$, generated from the output layer in the subsidiary branch.

$$pk = \frac{exp(X_{11(k)})}{\sum_k exp(X_{11(k)})} \qquad (6)$$

$$psk = \frac{exp(S_{8(k)})}{\sum_k exp(S_{8(k)})} \qquad (7)$$

Formula (8) [19] illustrates the loss calculated by the master branch by computing on probabilities initialized in the softmax. The auxiliary branch loss is calculated using formula (9) [19]. This loss includes weights from the auxiliary branch and the early convolutional layers from the master branch.

$$L_0(W_{main}) = -\sum_{k=1}^{K} yk \ln pk \qquad (8)$$

$$L_s(W_{main}, W_{branch}) = -\sum_{k=1}^{K} yk \ln psk \qquad (9)$$

Loss from the master and auxiliary branches can be calculated using formula (10) [19]. Formula (10) calculates a weighted sum as the master branch is exposed to more weight than the subsidiary branch. In order to manage the value of the subsidiary branch as a regularization parameter, we use the word $α_t$. This word degenerates over sequential iterations as illustrated in formula (11) [19].

$$L_s(W_{main}, W_{branch}) = L_0(W_{main}) + α_t L_s(W_{main}, W_{branch}) \qquad (10)$$
$$α_t = α_t * (1 – t/N) \qquad (11)$$

Formula (12) [22] uses shortcut connections from the residual learning [22] in ResSquCNDS.

$$y = F(x, \{W_{ij}\}) + x \qquad (12)$$

Following a deep study of the Squeeze-CNDS neural network, we decided to attach residual learning connections [23] to only the master branch. The residual connections were attached to places with sequences of convolutional layers and no pooling in between. Consequently, this left us unable to attach residual connections to the subsidiary branch as it contains no sequences of convolutional layers. Figure 3 illustrates our architecture showing residual connections in the main branch. Initially, the residual connection links input of the (Fire2) to the output of (Fire3) as the element-wise addition links the output of (Pool2) to output of (Fire3). The kernel of (Fire1) is 128 and kernel of (Fire3) are 256. In order to make the kernels' output equal, we use element-wise addition. We also connect a convolutional layer of kernel size 256 between (Pool2) and the element-wise addition layers.

While the second residual connection is connected next (Pool3) and the shortcut connection exceeds two Fire module layers. This results in the residual connection being connected between the output of (Pool3) and the output of (Fire5). The (Fire3) has a kernel of size 256 and (Fire5) has a kernel of size 512. We insert a convolutional layer with a kernel of size 512 after (Pool3) but before the element-wise addition layer in order to adjust the size of kernels of (Pool3) and (Fire5). We add the subsidiary branch after the integration process between the output of (Pool2) and (Fire3). Ultimately, the last (3$^{rd}$) residual connection links the output of (Pool4) to the (Fire7). Even though, it is not necessary to place an additional adjustment layer after (Pool4) and the element-wise addition due to the size of the kernel for (Fire5) and (Fire7) being 512 in both cases. We add a convolutional layer with a kernel of size 512 after (Pool4) but before the element-wise addition layer in order to boost the feature mapping in the end of the CNN.

## IV. IMAGE DATASET DESCRIPTION

MIT Places365-Standard [34] is a very large-scale dataset created and maintained by MIT Computer Science and Artificial Intelligence Laboratory. MIT Places365-Standard [34] dataset is bigger than ImageNet (ILSVRC2016) [35] and SUN dataset [32]. MIT Places365-Standard [34] dataset has 1,803,460 training images while each class contains anywhere from 3,068 to 5,000 images. MIT Places365-Standard [34] dataset has 50 image classes as validation set and 900 images/class as a test set. MIT Places365-Standard [34] dataset is scene based, meaning it includes images labeled with a scene/place name. The goal of the MIT Places365-Standard [34] dataset is to assist the academic goals in the field of computer vision. Our experiments were conducted on the MIT Places365-Standard [34] dataset.

## V. Experimental Environment and Approach

To begin, we trained both our Residual-CNDS [45] and our Residual-Squeeze-CNDS from scratch. Residual-CNDS [45] contains eight convolutional layers with three residual connections in the main branch and one convolutional layer in the subsidiary branch in contrast to the Residual-Squeeze-CNDS, which has four convolutional layers and seven Fire modules with three residual connections in the main branch, and four convolutional layers in the subsidiary branch. To complete our work, we use Caffe [28], an open source deep learning framework from the Berkeley Vision and Learning Center. In conjunction with Caffe, we use NVIDIA DIGITS an open source deep learning GPU training system [33], which allows users to build and examine their artificial neural networks for object detection and image classification with real-time visualization. As for physical hardware, we operate on four NVIDIA GeForce GTX TITAN X GPUs and two Intel Xeon processors allowing us a total of 48/24 logical/physical cores and 256 GB on the hard disk.

For our training, validation and testing data sets, all images are resized to 256x256. Our preprocessing activities are concluded after a subtraction of the average pixel for each color channel of RGB color space. We set the batch size for the training phase to 256, while we set the batch size of the validation to 128. We set our epoch count to 50, and we set the learning rate to 0.01. Our learning rate will degrade 5x during training after every 10 epochs and the decay of the learning average to half of its previous values. Images were cropped to 227x227 in random areas before being fed into the first convolutional layer. Then, the weights of all layers are adapted from the Xavier distribution with a 0.01 standard deviation. The final convolutional layer which acts as our output layer has its weight adapted from the Gaussian distribution with a 0.01 standard deviation as well. Reflection is the only augmentation used.

Our Residual-CNDS model [45] took one day and 21 hours to converge with a total size of (14gb). For comparison, the new Residual-Squeeze-CNDS took only one day and fifteen hours and a total size of (1.73gb). This means, our Residual-Squeeze-CNDS model is 13.33% faster 87.64%smaller in size that the original Residual-CNDS [45]. Which we utilized above-mentioned setup in the training, which Table 2 gives the Top-1 and Top-5 accuracy of the models in the validation classification accuracy.

## VI. RESULTS AND DISCUSSION

In conclusion, this paper has gathered three popular methods (convolutional neural networks with deep supervision [19], residual learning [22] and the Squeeze technique [40]). We set out to examine whether residual connections can boost CNDS [19] network effectiveness while simultaneously making the network smaller and faster. To do this we adapted and modified the Fire module concept [40] and utilized our sold method of determining when and where to add residual connections. We found residual connections to be parameter free, so even after a trivial amount of computation for the collection process, the complexity of the network does not see a large increase. Additionally, the fire modules [40] aided in the reduction of our network size, training time and complexity of our network with only a small Top-1 and Top-5 accuracy loss. Table 2 shows that after training from scratch, out Residual-Squeeze-CNDS Top-1 outcome is 51.32 whereas the original Residual-CNDS [45]

TABLE (2)

COMPARISON OF THE TOP 1 & 5 VALIDATION CLASSIFICATION ACCURACY (%), DURATION AND SIZE BETWEEN RESIDUAL-CNDS [45] AND RESIDUAL-SQUEEZE-CNDS ON THE MIT PLACES 365-STANDARD DATASET [34]

| Network | Top-1 Validation | Top-5 Validation | Duration | Size |
|---|---|---|---|---|
| ResCNDS | 51.98 | 82.11 | 1 Day 21 Hour | 14 GB |
| ResSquCNDS | 51.32 | 81.34 | 1 Day 15 Hour | 1.73 GB |

Top-1 outcome was almost a similar 51.98 on the validation set in the MIT Places 365-Standard dataset [34], which the different is only 0.66 percent. Our Residual-Squeeze-CNDS Top-5 result of 81.34 also showed a similarity to our original Residual-CNDS [45] Top-5 result of 82.11on the validation set, which the deference is only 0.77 percent. On the other hand, the Residual-Squeeze-CNDS is 87.64% smaller in size and 13.33% faster in the training time than the Residual-CNDS [45].

## VII. Conclusion and Future Work

This paper proposed a Residual-Squeeze-CNDS network with adaptations and modification on the fire module [40] and our state of the art method of determining when and where to add residual connections [45]. The fire module [40] not only helped us to achieve a reduction in the overall size and training time of our network but also a reduction in the network's computation complexity with only a very minute loss in Top-1 and Top-5 accuracy. Our network was tested the very large MIT Places 365-Standard [34] dataset, which illustrates our improvements in size, time and complexity from our older Residual-CNDS model [45].

Future work will focus on the application of the techniques we've outlined in this paper for Residual-Squeeze to other highly-regarded networks including but not limited to VGG [6], ResNet [22] and Densely Connected Convolutional Networks [47]. We hope to achieve similar reductions in size and complexity while only suffering little to no loss in Top-1 and Top-5 accuracy.